\documentclass[runningheads]{llncs}

\usepackage[year=2026]{eccv}

\usepackage[letterpaper,margin=1in]{geometry}

\usepackage{eccvabbrv}

\usepackage{graphicx}
\usepackage{booktabs}
\usepackage{amsmath}
\usepackage{amssymb}
\usepackage{multirow}
\usepackage{subcaption}
\usepackage{wrapfig}

\usepackage[accsupp]{axessibility}  

\usepackage[ruled,lined,linesnumbered]{algorithm2e}
\SetKwInput{KwInput}{Input}
\SetKwInput{KwOutput}{Output}
\SetKwComment{Comment}{$\triangleright$~}{}
\SetKwFunction{GreedyCoreset}{GreedyCoreset}
\DontPrintSemicolon

\newcommand{\noteA}[1]{}
\newcommand{\noteB}[1]{}

\usepackage[breaklinks,colorlinks,citecolor=eccvblue]{hyperref}

\usepackage{orcidlink}

\begin{document}

\title{Rethinking Continual Anomaly Detection on the Edge: Benchmarking Under Realistic Industrial Conditions}

\titlerunning{Rethinking Continual Anomaly Detection}

\author{Chad Weatherly, Sen Lin}

\authorrunning{Weatherly, Lin}

\institute{University of Houston}

\maketitle

\begin{abstract}
Continual anomaly detection (CAD) addresses the need for industrial inspection systems to adapt to evolving production conditions, yet existing methods share three critical gaps: unrealistic evaluation, no systematic comparison, and no consideration of edge deployment constraints. We introduce a unified benchmark combining discrete-task evaluation on structural and logical anomalies, a novel continuous drift protocol, the first head-to-head comparison of all published CAD methods, and computational efficiency profiling on edge hardware. Our results reveal that existing CAD methods do not consistently outperform traditional approaches with simple experience replay. Thus motivated, we propose DINOSaur, a training-free method combining a frozen DINOv3 backbone with spatially-indexed coreset memory and neighborhood-restricted anomaly scoring. DINOSaur achieves zero forgetting by construction, outperforms all evaluated methods on every protocol except geometric drift (where all methods collapse to chance), and runs at sub-100\,ms inference on an NVIDIA Jetson Orin Nano, with on-device adaptation to new tasks in under 30 seconds.
\end{abstract}

\section{Introduction}
\label{sec:intro}

The rise of Industry 4.0 has brought increased automation to manufacturing, including the adoption of AI-driven quality control. In production environments, poor quality control can be extremely costly, and missed defects can result in warranty claims, product recalls, and significant reputational damage~\cite{dimitriou_automatic_2025}. Anomaly detection (AD), which identifies defective products using models trained primarily on normal samples, has emerged as a natural fit for automated visual inspection~\cite{bergmann_mvtec_2019} in industrial environments.

However, real-world manufacturing processes are never truly static. Environmental conditions fluctuate, tools wear and degrade, material batches vary, and production parameters drift over time. These factors cause the distribution of normal product appearances to shift continuously, posing a fundamental challenge for AD systems trained under the assumption of a fixed data distribution. When the normal distribution changes, a previously deployed AD model may either miss genuine defects or flag normal variations as anomalous, both of which carry significant economic consequences.
Continual learning (CL) offers a framework that enables models to adapt to new data while retaining knowledge of previously learned distributions~\cite{wang_comprehensive_2024}. Several recent works have proposed methods for continual anomaly detection (CAD), combining AD architectures with CL strategies to maintain performance as data evolves across sequential tasks~\cite{li_towards_2022,tang_incremental_2025,liu_unsupervised_2024}.

Despite this progress, we identify three critical gaps in the current CAD literature:

\textbf{(1) Computational feasibility ignored.} Industrial AD systems typically run on edge devices with limited compute, often requiring multiple inference passes per second to keep pace with production throughput. Recent CAD methods employ large-scale architectures built on vision-language models (VLMs), diffusion models, or large vision transformers, with no analysis of whether these models could feasibly be deployed in the constrained environments~\cite{wyatt_anoddpm_2022,li_one-for-more_2025,gu_anomalygpt_2024,zhou_anomalyclip_2023}. While recent work studies efficient anomaly detection on edge hardware~\cite{barusco_paste_2025,barusco_memory_2025}, it targets static or single-backbone replay settings. Our feasibility concern is with dedicated CAD methods, whose added machinery for sequential learning and task identification has not been examined for edge deployment.

\textbf{(2) Unrealistic and narrow evaluation.} Existing CAD benchmarks treat each product categories as distinct tasks, creating sharp and discrete boundaries (\eg, \textit{bottle} to \textit{cable}). In reality, industrial drift is gradual---the same product changes appearance over time due to process variations. No existing benchmark captures this continuous drift scenario. Moreover, evaluation is limited to \textit{structural} anomalies (surface defects), ignoring \textit{logical} anomalies (missing components, incorrect arrangements)~\cite{bergmann_beyond_2022}---a significant unsolved challenge that no CAD method has been evaluated on.

\textbf{(3) Lack of systematic comparison.} The three published CAD methods, i.e., DNE~\cite{li_towards_2022}, IUF~\cite{tang_incremental_2025}, and UCAD~\cite{liu_unsupervised_2024}, were each evaluated independently against different baselines and under different experimental conditions. The three methods have never been comprehensively evaluated against one another, making it impossible to understand their relative strengths and weaknesses, let alone their feasibility in real-world scenarios.

To address these gaps, this paper makes the following contributions:
\begin{itemize}
    \item We introduce a \textbf{unified evaluation framework} for CAD that, for the first time, combines a head-to-head comparison of all published CAD methods, a novel \textit{continuous drift protocol}, evaluation on \textit{logical anomalies}, and computational efficiency profiling on edge hardware.
    \item Our evaluation reveals that \textbf{dedicated CAD methods do not consistently outperform} traditional baselines with simple experience replay. Motivated by this finding, we propose \textbf{DINOSaur}\footnote{Code and data: \url{https://github.com/Continue-Edge-AI-Lab/Rethinking-Continual-AD}}, a training-free method that outperforms all evaluated approaches at a fraction of the computational cost, enabling practical edge deployment.
\end{itemize}


\section{Related Work}
\label{sec:related}

\noindent\textbf{Anomaly Detection in Industrial Settings.}
Industrial anomaly detection (IAD) methods broadly fall into two paradigms. \textit{Distribution-based methods}~\cite{roth_towards_2022,defard_padim_2021,reiss_panda_2021} model the distribution of normal-sample features and flag out-of-distribution test samples as anomalous, using techniques such as nearest-neighbor scoring, multivariate Gaussian fitting, or learned normality scores. \textit{Reconstruction-based methods}~\cite{zavrtanik_draem_2021,you_unified_2022} train encoder-decoder networks on normal images and use high reconstruction error as an anomaly signal. Unified or multi-class AD~\cite{you_unified_2022} handles multiple categories simultaneously but assumes access to all data at once, unlike CAD's sequential constraint.

\noindent\textbf{Large-Scale Models for Anomaly Detection.}
Recent work has applied large foundation models to AD, including LLM-based reasoning~\cite{gu_anomalygpt_2024}, vision-language prompting~\cite{zhou_anomalyclip_2023}, in-context learning~\cite{zhu_toward_2024}, and diffusion models~\cite{wyatt_anoddpm_2022,li_one-for-more_2025}. While these achieve strong benchmark results, they require billions of parameters and GPU-class hardware with inference latencies measured in seconds per image---fundamentally incompatible with edge deployment where throughput of several images per second is required. Moreover, these large models are not immune to catastrophic forgetting: adapting vision-language models such as CLIP to sequential tasks causes significant degradation in zero-shot transfer ability~\cite{zheng_preventing_2023}, and multimodal LLMs similarly fail to retain their pre-trained visual understanding after fine-tuning~\cite{zhai_investigating_2024}. This dual challenge---computational infeasibility \textit{and} susceptibility to forgetting---further motivates our focus on lightweight, training-free approaches for edge-deployed CAD.

\noindent\textbf{Continual Learning.}
CL addresses the problem of learning from a sequence of tasks $\{D_1, D_2, \ldots, D_T\}$, subject to the constraint that when training on task $t$, only dataset $D_t$ is accessible (data from $D_1, \ldots, D_{t-1}$ is unavailable). Each task is a unique set of data, $D_t = {\{x_t, y_t\}}$ for $t \in [1, 2, \ldots, T]$. Under the CL framework, we aim to build a model that can continuously learn new tasks with minimum \textit{Catastrophic Forgetting} of the knowledge learned from previous tasks~\cite{wang_comprehensive_2024}. 

Existing CL strategies fall into three broad families. \textit{Regularization-based methods} add penalty terms to the loss function that restrict weight updates in order to preserve the important knowledge for previous tasks. Elastic Weight Consolidation (EWC)~\cite{kirkpatrick_overcoming_2017} uses Fisher information to identify important parameters, Synaptic Intelligence (SI)~\cite{zenke_continual_2017} tracks parameter importance online, and Memory Aware Synapses (MAS)~\cite{ferrari_memory_2018} computes importance based on output sensitivity. \textit{Architecture-based methods} allocate dedicated parameters or sub-networks for different tasks, as in Progressive Neural Networks~\cite{rusu_progressive_2022}. 
\emph{Memory-based methods} store information about previous tasks and modify the new task learning to leverage this information, which can be further divided into \textit{rehearsal-based methods} and \textit{gradient projection based methods}. The former stores raw exemplars or feature representations from previous tasks and replay them during training on new tasks, such as Experience Replay~\cite{rolnick_experience_2019} and Dark Experience Replay~\cite{buzzega_dark_2020}. The latter \cite{saha2021gradient,lin2022trgp,lin_beyond_2022} stores gradient information of previous tasks in this memory and uses this to modify the gradient direction for the new task. Empirically, memory-based methods consistently achieve the strongest performance across CL benchmarks~\cite{wang_comprehensive_2024}, as they most directly address forgetting by maintaining access to past information.

\noindent\textbf{Continual Anomaly Detection.}
To date, three published methods have specifically addressed the problem of CAD in realistic industrial settings.
\textbf{DNE}~\cite{li_towards_2022} stores per-task embedding statistics from a Vision Transformer (ViT) backbone and scores anomalies via Mahalanobis distance at image-level only. \textbf{IUF}~\cite{tang_incremental_2025} uses a ViT encoder-decoder with regularized latent space separation and constrained gradient updates. \textbf{UCAD}~\cite{liu_unsupervised_2024} combines contrastive learning with SAM~\cite{kirillov_segment_2023} for texture decomposition and a task-specific knowledge bank. Detailed architectural descriptions of each method are provided in \cref{sec:supp_related}.

Several benchmarks for CAD have emerged very recently, but don't compare CAD methods directly. Bugarin~\etal~\cite{bugarin_unveiling_2024} introduce a pixel-level CAD benchmark on MVTec-AD, evaluating off-the-shelf AD methods (PatchCore, CFA, STFPM, EfficientAD, FastFlow, among others) wrapped with standard replay or memory-bank strategies. Pezz\`{e}~\etal~\cite{pezze_continual_2025} similarly benchmark existing AD methods under replay, contributing a compressed-replay strategy that reduces memory via super-resolution but with no new detection architecture. Continual-MEGA~\cite{lee_continual-mega_2025} offers a large-scale benchmark and evaluates zero-shot generalization; its evaluated methods are CAD methods and CLIP-based VLMs, though it proposes a CLIP-based baseline with mixture-of-expert adapters specifically to mitigate forgetting in large models. A common thread across all three is reliance on generic CL strategies, predominantly replay, applied to conventional AD backbones, without methods designed around the specific constraints of continual anomaly detection (e.g.,\@ non-stationary normality, unsupervised, or edge deployment). Closest to our edge focus, Barusco~\etal\ reduce the cost of static AD on edge hardware~\cite{barusco_paste_2025} and study continual VAD on the edge with compressed replay~\cite{barusco_memory_2025}, but both apply generic replay to a single static backbone and do not compare dedicated CAD methods, continuous drift, or logical anomalies. Our benchmark differs in several further respects: we are the first to include a \textit{continuous drift} protocol (rather than only discrete category shifts), the first to evaluate CAD on \textit{logical anomalies} via MVTec-LOCO, and the first to profile \textit{computational efficiency on actual edge hardware}.

\section{A Unified Benchmark for Continual Anomaly Detection}
\label{sec:benchmark}

\subsection{Motivation}
\label{sec:benchmark_motivation}

The central motivation for our benchmark is the disconnect between how CAD methods are currently evaluated and how industrial AD systems actually operate. In a real manufacturing environment, an inspection system monitors the same product over time. The product's visual appearance changes gradually as environmental conditions shift (temperature, humidity, lighting), tools degrade, material batches change, or production processes drift. These changes create a continuous evolution of the normal data distribution---not the sharp, categorical shifts simulated by existing IAD datasets.
For example, a factory inspecting magnetic tiles for surface defects will experience gradual sensor degradation, seasonal lighting shifts, and material batch variation---all of which alter what ``normal'' looks like, yet the system must continue detecting genuine defects throughout. This is fundamentally different from a system that must suddenly transition from inspecting bottles to inspecting cables.

Beyond the distribution drift problem, existing CAD evaluations are also narrow in scope. Industrial defects are not limited to surface-level structural anomalies; logical anomalies, such as missing components and incorrect arrangements, are equally prevalent and require fundamentally different detection strategies, yet no published CAD method has been evaluated on them.

Our benchmark therefore evaluates CAD methods along three axes corresponding to the identified gaps: (1) \textit{evaluation realism and scope}, covering both the standard discrete task protocol and a novel continuous drift protocol alongside logical anomaly evaluation via MVTec-LOCO; (2) \textit{systematic comparison} of all existing CAD methods against each other and against traditional AD baselines; and (3) \textit{computational efficiency profiling} for edge deployment feasibility.

\subsection{Discrete Task Protocol}
\label{sec:benchmark_discrete}

For our discrete task protocol, we used the standard datasets for both AD and CAD~\cite{xie_im-iad_2024}:

\textbf{MVTec-AD.} MVTec-AD~\cite{bergmann_mvtec_2019} contains 15 object categories spanning textures and objects, each treated as a separate task learned sequentially. Only normal samples are used during training (unsupervised setting), reflecting the practical reality that anomalous samples are scarce in manufacturing. This protocol serves as our baseline for direct comparison with existing methods.

\textbf{MVTec-LOCO.} MVTec-LOCO~\cite{bergmann_beyond_2022} contains 5 categories that include both structural and \textit{logical anomalies}---defects requiring global scene understanding. The sequential protocol mirrors MVTec-AD, with each category as a task.

\subsection{Continuous Drift Protocol}
\label{sec:benchmark_drift}

To simulate realistic industrial data drift, we design a protocol based on the Magnetic Tile Defects (MTD) dataset~\cite{huang_saliency_2020}. MTD contains a single product category (magnetic tiles) with multiple defect types (blowhole, crack, break, fray, uneven), making it ideal for studying within-product distribution shift rather than between-product category changes.

\paragraph{Drift Simulation via Progressive Augmentation.}
We construct a sequence of $T = 10$ tasks from the MTD dataset by applying image augmentations of progressively increasing intensity. Let $\mathcal{A}(\mathbf{x}, \alpha)$ denote an augmentation function applied to image $\mathbf{x}$ with intensity $\alpha$. For task $t \in \{1, \ldots, T\}$, the intensity is greater for each successive task:
\begin{equation}
    \alpha_t > \alpha_{t-1} ~ \forall ~  t > 1.
    \label{eq:intensity}
\end{equation}
The data for each task is:
\begin{equation}
    D_t = \{ \mathcal{A}(\mathbf{x}, \alpha_t) \mid \mathbf{x} \in D_{\text{MTD}} \},
    \label{eq:task_data}
\end{equation}
where $D_{\text{MTD}}$ is the original MTD dataset. As $t$ increases, images undergo more severe distortion, simulating gradual environmental or process drift.

\paragraph{Augmentation Types.}
We run three separate experimental tracks, each simulating a different category of real-world drift:
\begin{itemize}
    \item \textbf{Color distortion}: simulates lighting changes and material batch variation via progressive hue, saturation, brightness, and contrast shifts.
    \item \textbf{Gaussian blur}: simulates sensor degradation and focus drift via increasing kernel size and standard deviation.
    \item \textbf{Geometric distortion}: simulates orientation changes and camera misalignment via progressive perspective transforms, rotations, and warping.
\end{itemize}

\begin{figure}[tb]
  \centering
  \includegraphics[width=\linewidth]{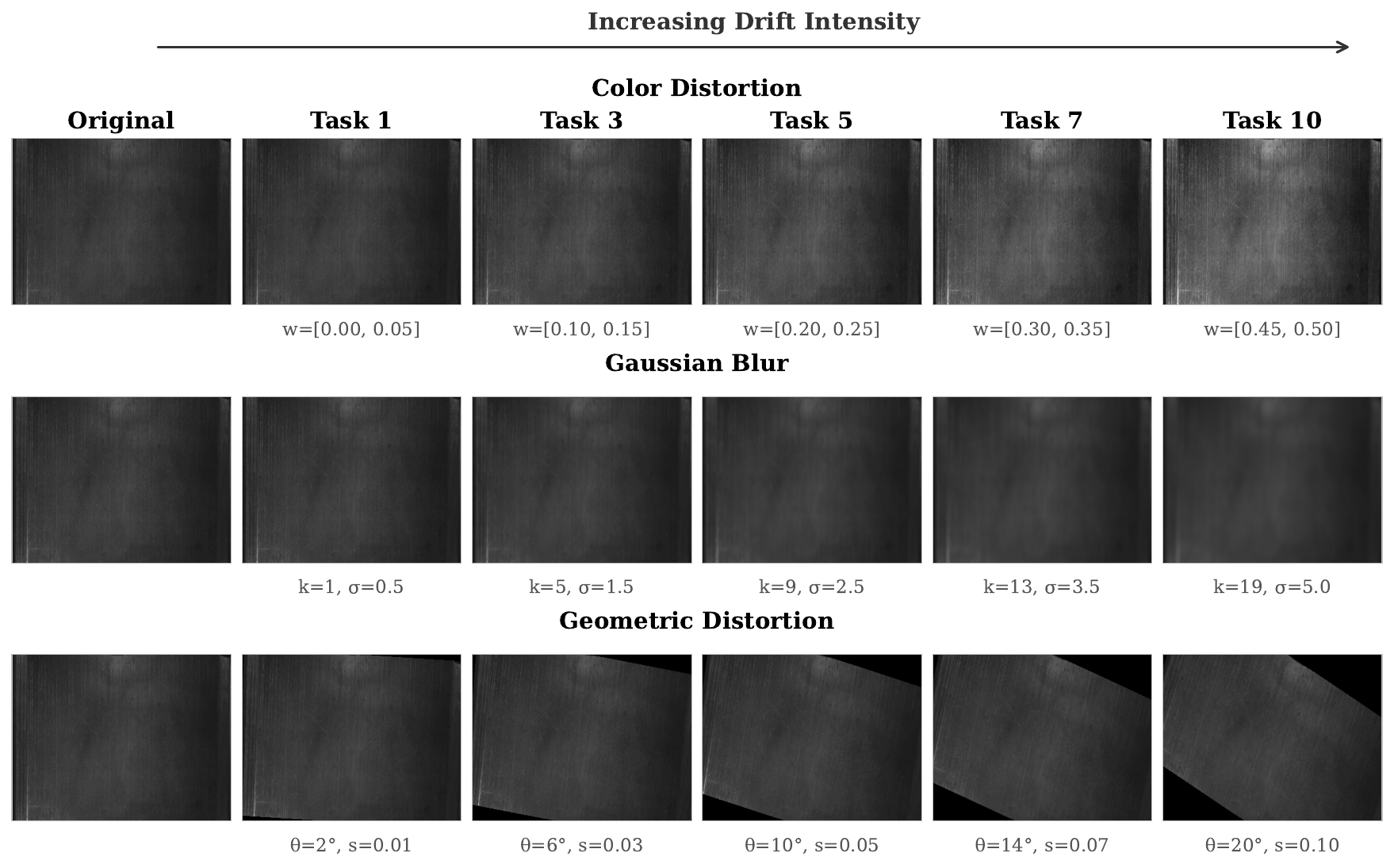}
   \vspace{-0.3cm}
  \caption{Progressive augmentation on MTD across tasks for each drift type.
    \emph{Top:} Color distortion (brightness, contrast, saturation). \emph{Middle:} Gaussian blur (increasing kernel size and $\sigma$). \emph{Bottom:} Geometric distortion (rotation, translation, scale, shear).
    Intensity increases from left (Task~1, minimal distortion) to right (Task~10, maximum distortion). Defect masks are unchanged under blur and color augmentation, and are transformed identically to the image under geometric augmentation.}
    \vspace{-0.5cm}
  \label{fig:augmentation_examples}
\end{figure}

Each augmentation type maps to documented sources of variation in industrial environments~\cite{faber_lifelong_2024}, providing interpretable and physically motivated drift scenarios rather than arbitrary distribution shifts. We empirically evaluated the intensity ranges to ensure that the magnetic tile objects were still visible. The exact parameter values for all 10 intensity levels of each drift type are provided in \cref{sec:supp_mtd} and illustrated in \Cref{fig:augmentation_examples}. 

\subsection{Evaluation Metrics}
\label{sec:benchmark_metrics}

Image-level anomaly scores are obtained by taking the maximum patch-level score. We evaluate all methods using the following metrics:

\textbf{Image-level AUROC} (Area Under the ROC Curve) measures threshold-independent discrimination between normal and anomalous images. We report the final AUROC averaged across all tasks after training on the last task.

\textbf{Image-level Accuracy} provides a threshold-dependent measure of classification performance on both normal and anomalous samples. 

\textbf{Recall (Sensitivity)} specifically measures the fraction of anomalous samples correctly identified, which is critical in industrial settings where missed defects (false negatives) carry higher costs than false positives. 
Accuracy and recall use a fixed threshold (97.5th percentile of training scores); details on threshold selection are in \cref{sec:supp_training}. Since AUROC is threshold-independent, it remains our primary metric.

\textbf{Pixel-level AUROC} measures localization of the defect region, complementing image-level detection. In distribution-based methods like DINOSaur and PatchCore, pixel-level anomaly maps are derived directly from the same patch-level scores used for image-level prediction, where the image score is the maximum over patch scores. We therefore expect the two to be tightly coupled, and we verify this in \cref{sec:pixel_eval}: across the protocols where pixel scoring is meaningful (excluding geometric drift, where transform padding dominates) they correlate at $r{=}0.945$ and preserve the ranking of the top methods, confirming that they are corollaries. Image-level AUROC remains our primary metric, as it matches the binary pass/fail decision common in industrial inspection, while pixel-level AUROC shows that our conclusions are not an artifact of metric choice.

\textbf{Forgetting Measure.}
To quantify catastrophic forgetting, we adopt the standard Forgetting Measure (FM)~\cite{wang_comprehensive_2024}. Let $a_{t,j}$ denote the performance on task $j$ after the model has been trained on task $t$. The forgetting for task $j$ after training on the final task $T$ is:
\begin{equation}
    f_j = \max_{t \in \{1, \ldots, T-1\}} a_{t,j} - a_{T,j},
    \label{eq:forgetting_per_task}
\end{equation}
and the average Forgetting Measure is:
\begin{equation}
    \text{FM} = \frac{1}{T-1} \sum_{j=1}^{T-1} f_j.
    \label{eq:forgetting_measure}
\end{equation}
A positive FM indicates performance degradation (catastrophic forgetting), while an FM of zero indicates zero forgetting of previous tasks.

\textbf{Computational Efficiency.}
We report model size (total parameters), inference latency (ms/image), and peak memory usage, which directly determine edge deployment feasibility on hardware such as NVIDIA Jetson and Raspberry Pi.

\section{DINOSaur: DINO Spatial Anomaly Unsupervised Recognition}
\label{sec:dinosaur}

\subsection{Motivation}
\label{sec:dinosaur_motivation}

\begin{figure}[tb]
  \centering
  \includegraphics[width=0.78\linewidth]{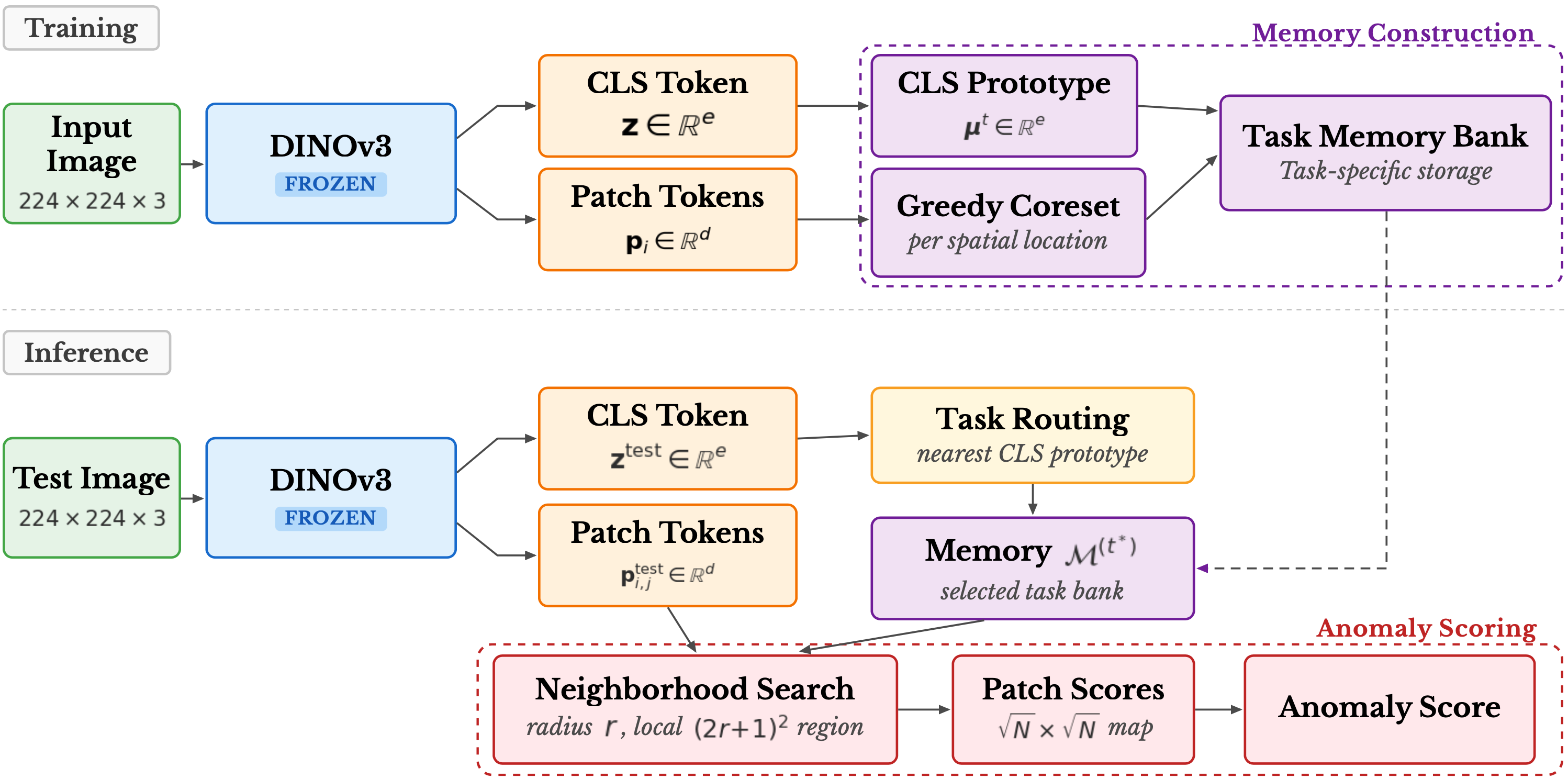}
  \caption{Overview of the DINOSaur architecture. \emph{Top:} During training, a frozen DINOv3 ViT-S/16 extracts CLS and patch tokens, which are stored in a task-specific memory bank via CLS prototyping and greedy coreset selection. \emph{Bottom:} During inference, the test image's CLS token is compared to stored prototypes for task routing, and patch tokens are scored against the selected memory bank using neighborhood-restricted nearest-neighbor search.}
  \label{fig:dinosaur_architecture}
  \vspace{-0.5cm}
\end{figure}

Existing CAD methods employ increasingly complex architectures---UCAD incorporates SAM~\cite{kirillov_segment_2023}, IUF trains a full ViT encoder-decoder, and DNE requires task-specific distribution estimation---yet, as we demonstrate in \cref{sec:experiments}, this complexity does not consistently translate into superior performance. Memory-based CL mechanisms consistently achieve the strongest results~\cite{wang_comprehensive_2024}, and in industrial settings long-term storage is rarely a constraint even on edge devices. We therefore hypothesize that a powerful frozen feature extractor paired with a non-parametric memory bank is sufficient: it yields competitive performance with zero forgetting by construction, since no model weights are updated. Thus motivated, we propose DINOSaur: a minimal baseline combining a frozen feature extractor with spatially-structured memory and neighborhood-based anomaly scoring, designed for edge deployment. An overview is shown in Figure~\ref{fig:dinosaur_architecture}.

\subsection{Feature Extraction with DINOv3}
\label{sec:dinosaur_features}

DINOSaur uses a frozen DINOv3 ViT-S/16~\cite{simeoni_dinov3_2025} as its feature backbone---the smallest variant in the DINOv3 family, with only 21M parameters---to demonstrate that strong CAD performance does not require large backbones. Larger variants (ViT-B/16, ViT-L/16) would likely improve detection quality at the cost of higher inference latency and memory usage; we deliberately select the smallest backbone to establish a lower bound representative of edge deployment constraints. We also chose to use the ViT version, as the CLS Token is necessary for task identification and routing. DINOv3 is a self-supervised vision transformer trained with student-teacher distillation and Gram anchoring, producing both a global CLS token and spatially distinctive patch-level features, considered current state of the art in producing embedding vectors with hierarchical conceptual understanding.

Given an input image $\mathbf{x} \in \mathbb{R}^{H \times W \times 3}$, the frozen DINOv3 encoder $\Phi$ produces a set of patch-level feature vectors:
\begin{equation}
    \{\mathbf{p}_1, \mathbf{p}_2, \ldots, \mathbf{p}_N\} = \Phi(\mathbf{x}), \quad \mathbf{p}_i \in \mathbb{R}^d,
    \label{eq:feature_extraction}
\end{equation}
where $N = (H/s) \times (W/s)$ is the number of patches determined by the patch size $s$, and $d$ is the feature dimension. We extract features from the final layer of the transformer, as these features contain descriptive, hierarchical and semantic understanding of the image patch.

Using a frozen backbone eliminates weight updates---the primary source of catastrophic forgetting---while DINOv3's self-supervised features provide strong semantic representations without task-specific adaptation.

\subsection{Spatially-Indexed Memory Bank}
\label{sec:dinosaur_coreset}

Rather than aggregating all patch features into a single flat memory bank, DINOSaur maintains a novel \textit{spatially-indexed} memory where each patch location has its own independent coreset. Given $D$ training images from the current task, we collect, for each spatial location $(i, j)$ in the $\sqrt{N} \times \sqrt{N}$ patch grid, the set of all feature vectors observed at that position across the task training set:
\begin{equation}
    \mathcal{P}_{i,j} = \{ \mathbf{p}_{i,j}^{(k)} \mid k = 1, \ldots, D \}, \quad |\mathcal{P}_{i,j}| = D.
    \label{eq:location_pool}
\end{equation}
We then apply greedy coreset selection~\cite{roth_towards_2022} independently at each location, iteratively choosing the point farthest from the current coreset, ensuring we have a representative set of location-specific feature vectors while also minimizing our memory bank size. This yields a per-location coreset $\mathcal{C}_{i,j}$ of size $M = \max(20,\; \lfloor D \cdot \rho \rfloor)$, where $\rho$ is the coreset ratio, a hyperparameter in the range $(0, 1]$. The full memory bank for a task is the spatially-structured tensor $\mathcal{M} \in \mathbb{R}^{\sqrt{N} \times \sqrt{N} \times M \times e}$, where $e$ is the embedding dimension of the feature vectors. This structure preserves the spatial layout of the image in the memory bank, which is essential for the neighborhood-based scoring described below.

\subsection{Neighborhood-Restricted Anomaly Scoring}
\label{sec:dinosaur_scoring}

During inference, a test image must first be routed to the appropriate task-specific memory bank (described in \cref{sec:dinosaur_cl}). Given the selected memory bank $\mathcal{M}$, we compute the anomaly score for each test patch by comparing it only to memory features within a local spatial neighborhood. Given a test patch at location $(i, j)$ with feature vector $\mathbf{p}_{i,j}^{\text{test}}$, we define the neighborhood memory as the union of patch coresets within a radius $r$:
\begin{equation}
    \mathcal{N}_{i,j}^{r} = \bigcup_{\substack{i' \in [i-r, i+r] \\ j' \in [j-r, j+r]}} \mathcal{C}_{i',j'}.
    \label{eq:neighborhood}
\end{equation}
The patch-level anomaly score is then:
\begin{equation}
    s_{i,j} = \min_{\mathbf{c} \in \mathcal{N}_{i,j}^{r}} \| \mathbf{p}_{i,j}^{\text{test}} - \mathbf{c} \|_2,
    \label{eq:patch_score}
\end{equation}
and the image-level score is the maximum over all patch locations:
\begin{equation}
    S(\mathbf{x}_{\text{test}}) = \max_{(i,j)} \; s_{i,j}.
    \label{eq:image_score}
\end{equation}

This neighborhood restriction enforces \textit{spatial consistency}---anomalies are detected based on what \textit{should} appear at a given location---and reduces computation from $N \cdot M$ to at most $(2r+1)^2 \cdot M$ comparisons per patch, a $4\times$ reduction for our default $r=3$ on a $14 \times 14$ grid.

\subsection{Continual Learning via Task-Specific Memory}
\label{sec:dinosaur_cl}

DINOSaur handles continual learning through two complementary mechanisms: \textbf{task-specific memory banks} and \textbf{CLS token prototypes} for unsupervised task identification.

\textbf{1) Task-specific memory.} Rather than accumulating all features into a single growing bank, DINOSaur maintains a separate, spatially-indexed memory bank $\mathcal{M}^{(t)}$ for each task $t$. When a new task arrives, its normal training images are processed through the frozen encoder, and a new per-location coreset is constructed and stored independently for that task, giving us a smaller memory footprint by using an optimal set of embedding vectors. Since the feature extractor $\Phi$ is never updated, the number of tasks able to be processed is limited mainly by memory and storage constraints, not model architecture.

\textbf{2) CLS token prototypes.} DINOv3 also produces a global CLS token $\mathbf{z} \in \mathbb{R}^e$ per image. During training on task $t$, we compute the mean CLS token $\boldsymbol{\mu}^{t}$ across all training images as a compact task prototype. At test time, unsupervised task routing selects the closest prototype:
\begin{equation}
    t^* = \arg\min_{t} \| \mathbf{z}^{\text{test}} - \boldsymbol{\mu}^{t} \|_2,
    \label{eq:task_id}
\end{equation}
and the selected task's memory bank $\mathcal{M}^{(t^*)}$ is used for scoring. This requires no task labels at inference---only a single distance computation per stored task.

\section{Experiments}
\label{sec:experiments}

\subsection{Experimental Setup}
\label{sec:exp_setup}

\subsubsection{Evaluated Methods.}
We evaluate the following methods:
\begin{itemize}
    \item \textbf{CAD methods}: DNE~\cite{li_towards_2022}, IUF~\cite{tang_incremental_2025}, and UCAD~\cite{liu_unsupervised_2024}---the three published continual anomaly detection methods.
    \item \textbf{Traditional AD baselines with replay}: PatchCore~\cite{roth_towards_2022} and EfficientAD~\cite{batzner_efficientad_2024}---augmented with experience replay~\cite{rolnick_experience_2019}. Replay budget and augmentation strategy details are in \cref{sec:supp_replay}.
    \item \textbf{DINOSaur} (ours)---the proposed baseline described in \cref{sec:dinosaur}.
\end{itemize}

For all methods, we use official implementations where available and faithfully reimplement otherwise, matching reported hyperparameters. All methods are trained for 300 epochs per task.

\subsubsection{Training Protocol.}
All experiments use the \textit{unsupervised} setting, where only normal samples are available during training, reflecting real industrial scenarios. Models are trained sequentially on each task without access to previous task data (the standard CL constraint), except for the generous replay buffer provided to PatchCore and EfficientAD. Task orderings for all protocols are provided in \cref{sec:supp_datasets}. Training is conducted on a single NVIDIA RTX 4090. Edge inference profiling is performed on an NVIDIA Jetson Orin Nano and a Raspberry Pi~5, as detailed in \cref{sec:exp_efficiency}.

\subsection{Main Results}
\label{sec:exp_discrete}

\Cref{tab:discrete_results} presents the image-level AUROC, Accuracy, Recall, and Forgetting Measure (FM) for all evaluated methods on MVTec-AD and MVTec-LOCO. Key findings:
\begin{itemize}
    \item DINOSaur and PatchCore with replay dominate on both protocols; dedicated CAD methods fall short.
    \item DNE and IUF collapse to predicting ``normal'' (zero recall on MVTec-AD), revealing fundamental architectural limitations in the unsupervised setting.
    \item Among dedicated CAD methods, only UCAD demonstrates meaningful learning, yet lags PatchCore by over 12 percentage points in AUROC on MVTec-AD.
    \item DINOSaur and UCAD achieve zero forgetting; PatchCore's modest forgetting shows that replay alone mitigates catastrophic forgetting.
\end{itemize}

While DINOSaur and PatchCore both occupy the top tier, the two methods differ in fundamental ways. PatchCore aggregates all patch features into a single flat memory bank and scores test patches via global nearest-neighbor search, relying on experience replay (100 samples per previous task) to mitigate forgetting when new tasks arrive. DINOSaur, by contrast, requires no replay, and its spatially-indexed memory with neighborhood-restricted scoring (\cref{sec:dinosaur_scoring}) enforces spatial consistency that a flat bank cannot. This architectural difference translates directly to edge performance: DINOSaur is 23\% faster than PatchCore on the Jetson (91.6 vs.\ 119\,ms; \cref{tab:efficiency}) while achieving higher AUROC on every protocol. Moreover, DINOSaur's performance is tunable via the coreset ratio $\rho$: on smaller or less variable tasks, reducing $\rho$ below the default 10\% would further decrease storage and latency with only modest AUROC loss (\cref{tab:hyperparam}), giving practitioners an explicit efficiency--accuracy knob. We note that our PatchCore${}^*$ baseline is mechanistically equivalent to PatchcoreCL~\cite{barusco_patchcorecl_2025}: both maintain persistent feature access without weight updates and use task-structured retrieval. Beyond the spatial indexing and neighborhood-restricted scoring above, DINOSaur further differs from both through explicit CLS-prototype task routing.

\begin{table*}[tb]
  \caption{Discrete-Task Results: Image-level AUROC ($\uparrow$), Accuracy ($\uparrow$), Recall ($\uparrow$), and Forgetting Measure ($\downarrow$) on MVTec-AD and MVTec-LOCO, averaged across all tasks. Methods marked with ${}^*$ are augmented with experience replay. Best results per column are in \textbf{bold}.}
  \label{tab:discrete_results}
  \centering
  \footnotesize
  \setlength{\tabcolsep}{3.5pt}
  \vspace{-0.4cm}
  \resizebox{12.2cm}{!}{%
  \begin{tabular}{@{}lcccccccc@{}}
    \toprule
    & \multicolumn{4}{c}{MVTec-AD} & \multicolumn{4}{c}{MVTec-LOCO} \\
    \cmidrule(lr){2-5} \cmidrule(lr){6-9}
    Method & AUROC & Acc & Recall & FM & AUROC & Acc & Recall & FM \\
    \midrule
    DNE~\cite{li_towards_2022}                            & 0.500 & 0.277 & 0.000 & 0.012 & 0.500 & 0.368 & 0.000 & \textbf{$-$0.022} \\
    IUF~\cite{tang_incremental_2025}                      & 0.508 & 0.300 & 0.074 & \textbf{$-$0.066} & 0.572 & 0.436 & 0.234 & 0.002 \\
    UCAD~\cite{liu_unsupervised_2024}                     & 0.798 & 0.635 & 0.552 & 0.000 & 0.629 & 0.537 & 0.389 & 0.000 \\
    \cmidrule(l){2-9}
    PatchCore${}^*$~\cite{roth_towards_2022}              & 0.925 & 0.808 & 0.804 & 0.023 & 0.743 & 0.655 & 0.521 & 0.038 \\
    EfficientAD${}^*$~\cite{batzner_efficientad_2024}     & 0.529 & 0.310 & 0.071 & 0.051 & 0.485 & 0.384 & 0.053 & 0.025 \\
    \cmidrule(l){2-9}
    DINOSaur (ours)                                       & \textbf{0.968} & \textbf{0.829} & \textbf{0.991} & 0.000 & \textbf{0.768} & \textbf{0.709} & \textbf{0.845} & 0.000 \\
    \bottomrule
  \end{tabular}%
  }
  \vspace{-0.4cm}
\end{table*}

The continuous drift results (\cref{tab:continuous_results}) reveal distinct patterns across drift types. On color and blur drift, DINOSaur and PatchCore clearly separate from dedicated CAD methods, confirming that strong features matter more than specialized CL mechanisms under continuous shift. Notably, continuous drift is not uniformly easier than discrete task shifts: DINOSaur and PatchCore achieve comparable or lower AUROC on MTD color and blur drift than on MVTec-AD, while the weaker CAD methods (DNE, IUF) show similar near-chance performance regardless of protocol, suggesting their failures are fundamental architectural limitations rather than protocol-specific artifacts.

Among drift types, geometric distortion is uniquely challenging---all methods collapse near chance. This is because geometric transformations fundamentally alter the spatial relationships between patch features, disrupting the very structure that distribution-based methods rely on. Color and blur distortions, by contrast, modify appearance while preserving spatial structure, making them far more amenable to patch-based detection. This distinction has practical implications: industrial scenarios involving mechanical deformation or perspective changes (\eg, conveyor belt misalignment, camera drift) may require fundamentally different representations than those sufficient for photometric variations. DINOSaur maintains near-zero forgetting across all drift types (FM\,$\leq$\,0.013), and on color drift PatchCore actually exhibits negative FM ($-$0.043), indicating backward transfer---continued training on shifted data \textit{improves} detection on earlier tasks.

\begin{table*}[tb]
  \caption{Continuous Drift Results: Image-level AUROC ($\uparrow$), Accuracy ($\uparrow$), Recall ($\uparrow$), and Forgetting Measure ($\downarrow$) on MTD-Color, MTD-Geo, and MTD-Blur, averaged across all tasks. Methods marked with ${}^*$ are augmented with experience replay. Best results per column are in \textbf{bold}.}
  \label{tab:continuous_results}
  \centering
  \vspace{-0.4cm}
  \resizebox{12.2cm}{!}{%
  \begin{tabular}{@{}lcccccccccccc@{}}
    \toprule
    & \multicolumn{4}{c}{MTD-Color} & \multicolumn{4}{c}{MTD-Geo} & \multicolumn{4}{c}{MTD-Blur} \\
    \cmidrule(lr){2-5} \cmidrule(lr){6-9} \cmidrule(lr){10-13}
    Method & AUROC & Acc & Recall & FM & AUROC & Acc & Recall & FM & AUROC & Acc & Recall & FM \\
    \midrule
    DNE~\cite{li_towards_2022}                            & 0.520 & 0.529 & 0.308 & 0.021 & 0.480 & 0.517 & \textbf{0.342} & 0.032 & 0.555 & 0.573 & 0.062 & $-$0.003 \\
    IUF~\cite{tang_incremental_2025}                      & 0.521 & 0.532 & 0.294 & $-$0.008 & \textbf{0.525} & \textbf{0.588} & 0.048 & 0.002 & 0.516 & 0.567 & 0.042 & 0.001 \\
    UCAD~\cite{liu_unsupervised_2024}                     & 0.566 & 0.591 & 0.141 & 0.030 & 0.480 & 0.548 & 0.107 & 0.007 & 0.582 & 0.586 & 0.137 & 0.003 \\
    \cmidrule(l){2-13}
    PatchCore${}^*$~\cite{roth_towards_2022}              & 0.836 & 0.710 & 0.530 & \textbf{$-$0.043} & 0.509 & 0.564 & 0.042 & \textbf{$-$0.008} & 0.832 & 0.751 & \textbf{0.685} & \textbf{$-$0.022} \\
    EfficientAD${}^*$~\cite{batzner_efficientad_2024}     & 0.565 & 0.573 & 0.080 & $-$0.007 & 0.510 & 0.567 & 0.050 & $-$0.004 & 0.566 & 0.564 & 0.051 & $-$0.010 \\
    \cmidrule(l){2-13}
    DINOSaur (ours)                                       & \textbf{0.928} & \textbf{0.836} & \textbf{0.861} & 0.000 & 0.519 & 0.567 & 0.050 & 0.013 & \textbf{0.891} & \textbf{0.810} & 0.676 & $-$0.002 \\
    \bottomrule
  \end{tabular}%
  }
  \vspace{-0.2cm}
\end{table*}

\subsection{Pixel-level Evaluation}
\label{sec:pixel_eval}

\Cref{tab:pixel_results} reports image- and pixel-level AUROC side by side. Across the protocols where pixel scoring is meaningful (MVTec-AD, MVTec-LOCO, MTD-color, MTD-blur), the two correlate at $r{=}0.945$ and preserve the ranking of the top methods, confirming that pixel-level performance is largely a corollary of image-level detection. DINOSaur retains its pixel-level lead on all four. We omit geometric pixel-AUROC, where the affine padding dominates the pooled metric. DNE has no spatial map, so its pixel entries are undefined.

\begin{table*}[tb]
  \caption{Image- vs.\ pixel-level AUROC ($\uparrow$), averaged across all tasks. Image-level numbers are reproduced from \cref{tab:discrete_results,tab:continuous_results}. DNE has no spatial map (``---''). Under geometric drift, pixel-level scoring is unreliable because the affine padding dominates the pooled metric, so we omit it (``---''). Best pixel-level result per protocol is in \textbf{bold}.}
  \label{tab:pixel_results}
  \centering
  \scriptsize
  \setlength{\tabcolsep}{3pt}
  \vspace{-0.4cm}
  \resizebox{12.2cm}{!}{%
  \begin{tabular}{@{}lcccccccccc@{}}
    \toprule
    & \multicolumn{2}{c}{MVTec-AD} & \multicolumn{2}{c}{MVTec-LOCO} & \multicolumn{2}{c}{MTD-Color} & \multicolumn{2}{c}{MTD-Geo} & \multicolumn{2}{c}{MTD-Blur} \\
    \cmidrule(lr){2-3} \cmidrule(lr){4-5} \cmidrule(lr){6-7} \cmidrule(lr){8-9} \cmidrule(lr){10-11}
    Method & img & pix & img & pix & img & pix & img & pix & img & pix \\
    \midrule
    DNE~\cite{li_towards_2022}                        & 0.500 & --- & 0.500 & --- & 0.520 & --- & 0.480 & --- & 0.555 & --- \\
    IUF~\cite{tang_incremental_2025}                  & 0.508 & 0.525 & 0.572 & 0.606 & 0.521 & 0.557 & 0.525 & --- & 0.516 & 0.585 \\
    UCAD~\cite{liu_unsupervised_2024}                 & 0.798 & 0.822 & 0.629 & 0.651 & 0.566 & 0.564 & 0.480 & --- & 0.582 & 0.537 \\
    \cmidrule(l){2-11}
    PatchCore${}^*$~\cite{roth_towards_2022}          & 0.925 & 0.945 & 0.743 & 0.742 & 0.836 & 0.770 & 0.509 & --- & 0.832 & 0.730 \\
    EfficientAD${}^*$~\cite{batzner_efficientad_2024} & 0.529 & 0.402 & 0.485 & 0.429 & 0.565 & 0.599 & 0.510 & --- & 0.566 & 0.636 \\
    \cmidrule(l){2-11}
    DINOSaur (ours)                                   & \textbf{0.968} & \textbf{0.947} & \textbf{0.768} & \textbf{0.810} & \textbf{0.928} & \textbf{0.875} & 0.519 & --- & \textbf{0.891} & \textbf{0.857} \\
    \bottomrule
  \end{tabular}%
  }
  \vspace{-0.4cm}
\end{table*}

\subsection{Computational Efficiency}
\label{sec:exp_efficiency}

\begin{table*}[tb]
  \caption{Computational efficiency on two edge platforms (NVIDIA Jetson Orin Nano, Raspberry Pi 5). Latency is mean $\pm$ std over 30 single-image runs. Storage is measured on-device per task; ``---'' marks weight-update methods with no task-indexed storage (DINOSaur storage shown for MVTec-AD). IUF exceeded the Jetson's memory (OOM). FPS = frames per second; ${}^*$ = experience replay. Best latency per platform in \textbf{bold}.}
  \label{tab:efficiency}
  \centering
  \vspace{-0.4cm}
  \resizebox{12.2cm}{!}{%
  \begin{tabular}{@{}lcccccc@{}}
    \toprule
    & & & \multicolumn{2}{c}{Jetson Orin Nano} & \multicolumn{2}{c}{Raspberry Pi 5} \\
    \cmidrule(lr){4-5} \cmidrule(lr){6-7}
    Method & Params (M) & Storage/Task (MB) & Inf.\ (ms) & FPS & Inf.\ (ms) & FPS \\
    \midrule
    DNE~\cite{li_towards_2022}                        & 86.6  & 2.3            & $60.7 \pm 4.9$     & 16.5 & $\mathbf{560.0 \pm 6.1}$    & \textbf{1.8}  \\
    IUF~\cite{tang_incremental_2025}                  & 631.1 & ---            & OOM                 & ---  & $3668.6 \pm 69.6$  & 0.3  \\
    UCAD~\cite{liu_unsupervised_2024}                 & 36.0  & 1.2            & $956.5 \pm 7.3$    & 1.0  & $1058.0 \pm 15.4$  & 0.9  \\
    \midrule
    PatchCore${}^*$~\cite{roth_towards_2022}          & 9.3   & 5.0            & $119.0 \pm 2.4$    & 8.4  & $1074.5 \pm 7.9$   & 0.9  \\
    EfficientAD${}^*$~\cite{batzner_efficientad_2024} & 8.1   & ---            & $\mathbf{34.7 \pm 0.1}$ & \textbf{28.8} & $1318.2 \pm 4.3$ & 0.8  \\
    \midrule
    DINOSaur (ours)                                   & 21.6  & 7.1            & $91.6 \pm 9.6$     & 10.9 & $630.6 \pm 10.2$   & 1.6  \\
    \bottomrule
  \end{tabular}%
  }
  \vspace{-0.6cm}
\end{table*}

\cref{tab:efficiency} compares the computational requirements of all evaluated methods on two representative edge platforms. Industrial inspection systems typically require processing multiple images per second on hardware with limited GPU memory; a method that cannot meet these constraints is not a viable CAD solution regardless of its detection metrics.

Among the dedicated CAD methods, IUF exceeds the Jetson's GPU memory entirely and requires 3.7 seconds per image on the Raspberry Pi. UCAD's reliance on SAM~\cite{kirillov_segment_2023} results in nearly 1 second per image on both devices. DNE is the most efficient method overall---fastest on the Raspberry Pi (560\,ms, 1.8 FPS) and competitive on the Jetson (60.7\,ms)---but achieves near-chance AUROC, making its speed advantage moot in practice.

Among methods with meaningful detection, DINOSaur achieves sub-100\,ms inference on the Jetson (91.6\,ms, 10.9 FPS), faster than PatchCore (119\,ms). EfficientAD is faster (34.7\,ms) but collapses in detection under continual learning. On the CPU-only Raspberry Pi~5, DINOSaur (630.6\,ms) is second-fastest after DNE. Per-task storage (7.1\,MB on MVTec-AD) scales with $\rho$ and training set size; details are in \cref{sec:supp_storage}.

\textbf{On-Device Training Feasibility.}
Training cost matters as much as inference for edge deployment. Gradient-based methods (DNE, IUF, UCAD, EfficientAD) require forward and backward passes over hundreds of epochs per task, with memory far exceeding inference. DINOSaur instead adapts to a new task with a single forward pass per image plus coreset selection, enabling autonomous on-device adaptation without cloud connectivity.

\subsection{Analysis and Discussion}
\label{sec:exp_analysis}
On the four protocols with meaningful signal, no dedicated CAD method achieves the top AUROC; DINOSaur leads all four, followed by PatchCore with replay. Under geometric drift every method, dedicated or not, sits at chance. This raises a fundamental question: are the performance gains reported in prior CAD papers genuine algorithmic advances, or artifacts of experimental protocol? We identify the \textit{unsupervised} setting as a key factor. All of our experiments train on normal samples only, which is substantially harder than the supervised setting used in some prior evaluations of DNE and IUF, which use synthetic anomaly generation. The near-chance performance of these methods suggests that their architectures may rely on anomaly examples during training to learn meaningful decision boundaries---a luxury that real industrial deployments rarely afford. When this crutch is removed, dedicated CL mechanisms provide no measurable benefit over simple replay, and the added architectural complexity yields no return on investment for edge-constrained deployment.

\paragraph{Task Ordering Invariance.}
Because DINOSaur's backbone is frozen and each task receives an independent memory bank, its final performance is \textit{completely invariant} to task ordering---a structural guarantee no other evaluated method provides, since all others either update weights or replay across tasks. This eliminates a confound in CL evaluation and simplifies deployment. Extended analysis is in \cref{sec:supp_analysis}.

\begin{wraptable}{r}{5.5cm}
  \centering
  \caption{Hyperparameter sensitivity: average image-level AUROC across all five benchmark protocols for each coreset ratio ($\rho$) and neighborhood radius ($r$) combination. The selected configuration ($\rho{=}10\%$, $r{=}3$) is \textbf{bolded}.}
  \label{tab:hyperparam}
  \footnotesize
  \setlength{\tabcolsep}{6pt}
  \resizebox{5.5cm}{!}{%
  \begin{tabular}{@{}lccccc@{}}
    \toprule
    & \multicolumn{5}{c}{Neighborhood radius $r$} \\
    \cmidrule(l){2-6}
    Coreset ratio $\rho$ & 0 & 1 & 2 & 3 & 4 \\
    \midrule
    1\% & 0.771 & 0.773 & 0.773 & 0.780 & 0.778 \\
    2.5\% & 0.765 & 0.770 & 0.775 & 0.776 & 0.782 \\
    5\% & 0.771 & 0.775 & 0.777 & 0.781 & 0.785 \\
    10\% & 0.787 & 0.790 & 0.794 & \textbf{0.801} & 0.798 \\
    20\% & 0.803 & 0.816 & 0.819 & 0.816 & 0.822 \\
    \bottomrule
  \end{tabular}%
  }
  \vspace{-0.4cm}
\end{wraptable}
\paragraph{DINOSaur Hyperparameter Sensitivity.}
DINOSaur has two hyperparameters: the coreset ratio $\rho$ and the neighborhood radius $r$. We perform a grid search over $\rho \in \{1\%, 2.5\%, 5\%, 10\%, 20\%\}$ and $r \in \{0, 1, 2, 3, 4\}$, reporting the average AUROC across all five protocols in \cref{tab:hyperparam}. Performance increases monotonically with $\rho$ but with diminishing returns (1.5\% gain from 10\% to 20\%), while $r$ has a smaller but consistent effect. We select $\rho{=}10\%$ and $r{=}3$, balancing detection quality against edge efficiency (sub-100\,ms inference on the Jetson). The full per-dataset breakdown is in the appendix.

\section{Conclusion}
\label{sec:conclusion}

We introduced a unified CAD benchmark addressing three gaps in the literature, i.e., unrealistic evaluation, lack of systematic comparison, and neglect of edge constraints, and showed that existing CAD methods do not consistently outperform traditional AD baselines with simple experience replay. Motivated by this finding, we proposed DINOSaur, which combines a frozen DINOv3 backbone with spatially-indexed coreset memory and task identification/routing to achieve state-of-the-art performance with zero forgetting by construction across four of the five benchmark protocols (every method collapses to chance under geometric drift). Its training-free design enables on-device adaptation in under 30 seconds on an NVIDIA Jetson Orin Nano, making it practical for autonomous industrial deployment.

\paragraph{Future Work.} Three directions remain open. First, geometric drift collapses every method to chance; addressing it will likely require geometrically equivariant feature representations beyond what current frozen backbones provide. Second, DINOSaur's neighborhood scoring is inherently local, so logical anomalies that require holistic scene understanding would benefit from added global reasoning alongside the spatial memory. Finally, extending the benchmark to online streaming without explicit task boundaries would bring evaluation closer to real industrial deployment.


%
%
\bibliographystyle{splncs04}
\bibliography{main}

\clearpage
\appendix

This appendix provides supporting material referenced from the main text: extended method descriptions (\cref{sec:supp_related}), detailed training procedures (\cref{sec:supp_training}), per-task dataset specifications (\cref{sec:supp_datasets}), extended analysis (\cref{sec:supp_analysis}), and full algorithmic pseudocode for DINOSaur (\cref{sec:supp_algorithm}).

\section{Extended Related Work}
\label{sec:supp_related}

This section expands on the method descriptions provided in the main text (Section~2). We discuss large-scale models for anomaly detection in the main text (Section~2.1); here we provide additional architectural details for the three dedicated continual anomaly detection methods.

\subsection{Continual Anomaly Detection Methods}
\label{sec:supp_cad_methods}

\textbf{DNE} (Distribution of Normal Embeddings)~\cite{li_towards_2022} introduced the first framework for CAD. During training on each task, DNE extracts feature embeddings using a vision transformer (ViT-B/16) backbone and stores the mean vector, shrunk covariance matrix, and sample count of the embedding distribution per task. During inference, a global distribution is reconstructed from the stored per-task statistics, and anomaly scores are computed via Mahalanobis distance. DNE operates at image-level only and does not produce pixel-level segmentation maps. Notably, the feature embeddings are not regularized to be task-specific, meaning the stored statistics may not accurately capture task-distinct distributions as the number of tasks grows.

\textbf{IUF} (Incremental Unified Framework)~\cite{tang_incremental_2025} uses an encoder-decoder architecture with ViT components (embedding dimension 64, 4 attention heads, 4 layers), where the latent space is explicitly regularized to separate task-specific distributions into non-overlapping regions via a V-projection mechanism. Gradient updates are constrained to preserve important parameters for previous tasks through a weight regularization term ($\beta = 0.5$) that penalizes divergence from the previous task's parameters. The training objective combines reconstruction loss, discriminator loss, and a singular value penalty to encourage low-rank latent representations.

\textbf{UCAD} (Unsupervised Continual Anomaly Detection)~\cite{liu_unsupervised_2024} combines contrastive learning with a prompting mechanism for task identification and adaptation. It leverages the Segment Anything Model (SAM)~\cite{kirillov_segment_2023} to decompose image textures, then uses contrastive learning to store task-specific feature distributions in a knowledge bank for inference. UCAD maintains three memory components: a key memory (196 representative patches via farthest point sampling), a knowledge memory (196 patches via greedy coreset selection), and a prompt memory (learned task-specific prompts). Task identification at inference time uses the key memory to route test images to the appropriate knowledge bank.

\section{Training Procedures}
\label{sec:supp_training}

All experiments share a common training framework. We describe the shared configuration and per-method details below.

\subsection{Shared Configuration}
\label{sec:supp_shared_config}

All methods are trained sequentially on tasks in a fixed order (see \cref{sec:supp_datasets} for per-dataset task orderings). We use the Adam optimizer with a learning rate of $7.5 \times 10^{-4}$ and weight decay of $10^{-4}$ across all methods. The training batch size is 12 for all methods. Input images are resized to $224 \times 224$ pixels (except EfficientAD, which uses $256 \times 256$) and normalized with ImageNet statistics ($\mu = [0.485, 0.456, 0.406]$, $\sigma = [0.229, 0.224, 0.225]$). All training is performed on a single NVIDIA RTX 4090 GPU (24\,GB VRAM).

For methods that support experience replay (PatchCore and EfficientAD), 100 nominal training samples from previous tasks are replayed during each new task's training. No data augmentation is applied to replay samples. DINOSaur does not use experience replay; its frozen backbone and task-specific memory banks eliminate the need for replaying old data.

\subsection{Per-Method Training Details}
\label{sec:supp_per_method}

\cref{tab:supp_training} summarizes the key training parameters for each method.

\begin{table*}[tb]
  \caption{Training hyperparameters and key architectural details for all evaluated methods. ``Epochs'' indicates the number of full passes over the training data per task. Methods marked with $\dagger$ require only a single forward pass for feature extraction (no gradient-based optimization), and thus complete training in one epoch.}
  \label{tab:supp_training}
  \centering
  \setlength{\tabcolsep}{4pt}
  \resizebox{12.2cm}{!}{%
  \begin{tabular}{@{}lcccccc@{}}
    \toprule
    & DNE & IUF & UCAD & PatchCore${}^*$ & EfficientAD${}^*$ & DINOSaur \\
    \midrule
    Backbone & ViT-B/16 & Custom ViT & Custom ViT & ResNet-18 & PDN & DINOv3 ViT-S/16 \\
    Backbone frozen & No & No & Yes & Yes & Partially & Yes \\
    Embed.\ dim. & 768 & 64 & 768 & 512 & 384 & 384 \\
    Epochs/task & 300 & 300 & 300 & 1${}^\dagger$ & 300 & 1${}^\dagger$ \\
    Loss & CE & Composite & Contrastive & --- & MSE (3-part) & --- \\
    Replay & No & No & No & Yes (100) & Yes (100) & No \\
    Trainable params (M) & 85.98 & 631.07 & Prompt only & 0.79 & 5.38 & 0.00 \\
    \bottomrule
  \end{tabular}%
  }
\end{table*}

\subsubsection{DNE.}
DNE fine-tunes a ViT-B/16 backbone (86.57M total parameters) with a linear classification head (768 $\to$ 2) using cross-entropy loss over 300 epochs per task. The classification head is frozen after the first task, and subsequent tasks only update the backbone. After training each task, DNE stores per-task statistics: a 768-dimensional mean vector and a $768 \times 768$ shrunk covariance matrix (shrinkage $\alpha = 0.5$), yielding approximately 2.3\,MB of storage per task. At inference, a global distribution is reconstructed from all stored task statistics, and anomaly scores are computed via Mahalanobis distance. The anomaly threshold is set at the 97.5th percentile of training set anomaly scores.

\subsubsection{IUF.}
IUF trains an encoder-decoder architecture (631.07M parameters) with a composite loss function:
\begin{equation}
  \mathcal{L}_{\mathrm{IUF}} = \lambda_1 \mathcal{L}_{\mathrm{recon}} + \lambda_2 \mathcal{L}_{\mathrm{disc}} + \lambda_3 \mathcal{L}_{\mathrm{svd}},
\end{equation}
where $\lambda_1 = 1.0$, $\lambda_2 = 0.5$, and $\lambda_3 = 8.0$. The singular value loss $\mathcal{L}_{\mathrm{svd}}$ encourages low-rank latent representations by penalizing singular values beyond a cutoff $t = 1$. After each task, model weights are saved and a regularization term ($\beta = 0.5$) pulls parameters toward the previous task's weights during subsequent training. IUF operates only in the unsupervised setting; the supervised pathway was not used in our experiments. The anomaly threshold is set at the 97.5th percentile.

\subsubsection{UCAD.}
UCAD uses a frozen ViT backbone with a learnable prompt parameter of shape $(5, 768)$ trained via contrastive loss:
\begin{equation}
  \mathcal{L}_{\mathrm{UCAD}} = \lambda_\alpha \mathcal{L}_{\mathrm{neg}} - \lambda_\beta \mathcal{L}_{\mathrm{pos}},
\end{equation}
where $\lambda_\alpha = \lambda_\beta = 1.0$ and the loss components use cosine similarity between patch features. After training each task for 300 epochs, UCAD updates three memory banks: key memory (196 patches via farthest point sampling), knowledge memory (196 patches via greedy coreset), and the learned prompt. At inference, the 5 nearest neighbors in the knowledge bank determine patch-level anomaly scores, which are upsampled to the original image resolution. The anomaly threshold is set at the 97.5th percentile.

\subsubsection{PatchCore.}
PatchCore uses a frozen ResNet-18 backbone (9.33M parameters) and requires only a single epoch of feature extraction per task (no gradient updates). We deliberately select the smaller ResNet-18 variant to establish a lower bound representative of edge deployment, consistent with our choice of ViT-S/16 for DINOSaur. Features from intermediate layers are concatenated and projected to 512 dimensions via a random linear projection, then pooled with an adaptive average pooling layer (kernel size 3, stride 1, padding 1), producing a $28 \times 28$ spatial feature map per image. A greedy coreset of 5\% of the accumulated feature vectors is retained in the memory bank (minimum 500 samples). During inference, anomaly scores are computed as the distance to the nearest coreset entry, re-weighted by the 3 nearest neighbors of that match. The anomaly threshold is set at the 97.5th percentile.

\subsubsection{EfficientAD.}
EfficientAD trains a student Patch Description Network (PDN) and an autoencoder to distill knowledge from a pre-trained teacher PDN. The teacher is pre-trained on ImageNet with a separate optimizer (Adam, lr $= 10^{-3}$) for 60{,}000 iterations at batch size 16, using ResNet34 features normalized via Welford's running statistics (computed on 10{,}000 samples). The student and autoencoder are jointly trained for 300 epochs per task with a three-part MSE loss:
\begin{equation}
  \mathcal{L}_{\mathrm{EAD}} = \mathcal{L}_{\mathrm{ST}} + \mathcal{L}_{\mathrm{AE}} + \mathcal{L}_{\mathrm{STAE}},
\end{equation}
where $\mathcal{L}_{\mathrm{ST}}$ uses hard example mining ($p_{\mathrm{hard}} = 0.999$), $\mathcal{L}_{\mathrm{AE}}$ distills teacher features into the autoencoder, and $\mathcal{L}_{\mathrm{STAE}}$ aligns student and autoencoder outputs. The student outputs a 768-dimensional feature (two 384-dim halves), and the final anomaly map combines student-teacher and student-autoencoder discrepancies with equal weight (0.5 each). Input images are resized to $256 \times 256$ (student/teacher) and $512 \times 512$ (ResNet34 feature extractor). The anomaly threshold is set at the 97.5th percentile.

\subsubsection{DINOSaur.}
DINOSaur uses a completely frozen DINOv3 ViT-S/16 backbone (21.6M parameters, 0 trainable) and requires only a single epoch of feature extraction per task. The backbone produces 196 patch tokens of dimension 384 on a $14 \times 14$ spatial grid plus a 384-dimensional CLS token. Per-location greedy coreset selection retains $\rho = 10\%$ of patch features (minimum 20 per location), and the CLS token mean is stored as the task prototype. At inference, the test image's CLS token is compared against all stored prototypes for task routing, and patch anomaly scores are computed within a spatial neighborhood of radius $r = 3$ (yielding a $7 \times 7$ window). The image-level score is the maximum patch anomaly score. The anomaly threshold is set at the 97.5th percentile of training set scores. The coreset ratio $\rho$ and neighborhood radius $r$ were selected via grid search over $\rho \in \{1\%, 2.5\%, 5\%, 10\%, 20\%\}$ and $r \in \{0, 1, 2, 3, 4\}$; see the main text for details. Full algorithmic details are provided in \cref{sec:supp_algorithm}.

\subsubsection{Replay Strategy Details.}
\label{sec:supp_replay}

For PatchCore and EfficientAD, we employ a generous replay budget of 100 nominal training samples from previous tasks during each new task's training. This replay budget was selected to give replay-augmented baselines a strong chance at retaining prior knowledge. On discrete-task benchmarks (MVTec-AD and MVTec-LOCO), the typical category contains 200--400 training images, such that 100 samples represents 25--50\% of the original training set. No data augmentation is applied to replay samples. For the MTD continuous drift protocol, the augmentation parameter range is widened to encompass the distributions of previous task training data, providing traditional methods with access to ample past information and isolating the effect of dedicated CL mechanisms. DINOSaur does not use experience replay: because its backbone is frozen and each task receives an independent memory bank, there is no mechanism through which old data could influence the new task's memory, and no forgetting to mitigate.

\subsubsection{DINOSaur Per-Task Storage.}
\label{sec:supp_storage}
DINOSaur's per-task storage depends on the training set size $D$ and the coreset ratio $\rho$, since the memory bank stores $M = \max(20, \lfloor D \cdot \rho \rfloor)$ feature vectors of dimension 384 at each of the 196 spatial locations, plus a 384-dimensional CLS prototype. The theoretical per-task storage in bytes is $196 \times M \times 384 \times 4 + 384 \times 4$ (using 32-bit floats), yielding approximately 5.8\,MB for $M{=}20$. We report the measured storage throughout the paper.

\begin{table}[tb]
  \caption{DINOSaur task storage at $\rho = 10\%$ for each benchmark dataset. $D$ denotes the approximate number of training images per task. The minimum coreset size of 20 per location applies when $\lfloor D \cdot \rho \rfloor < 20$. Storage values are measured on-device and include serialization overhead.}
  \label{tab:supp_storage}
  \centering
  \setlength{\tabcolsep}{4pt}
  \resizebox{12.2cm}{!}{%
  \begin{tabular}{@{}lccccc@{}}
    \toprule
    Dataset & Tasks & $D$ (approx.) & $M$ per loc. & Total (MB) \\
    \midrule
    MVTec-AD & 15 & 60--391 & 20--39 & 106.8 \\
    MVTec-LOCO & 5 & 357--572 & 35--57 & 35.6 \\
    MTD (all drift types) & 10 & 443 & 44 & 71.2 \\
    \bottomrule
  \end{tabular}%
  }
\end{table}

At $\rho = 10\%$, most MVTec-AD categories (with $D < 200$) are dominated by the minimum coreset size of 20, while larger categories (e.g., hazelnut with $D{=}391$, yielding $M{=}39$) and the MTD dataset ($D{=}443$, $M{=}44$) exceed this minimum. Larger coreset ratios or larger datasets would increase per-task storage proportionally. In deployment scenarios with thousands of training images per task, storage would grow accordingly, and practitioners should consider the tradeoff between coreset ratio and available RAM on the target edge device.

\section{Dataset and Task Specifications}
\label{sec:supp_datasets}

\subsection{MVTec-AD}
\label{sec:supp_mvtec}

MVTec-AD~\cite{bergmann_mvtec_2019} comprises 15 industrial inspection categories spanning textures and objects. Each category is treated as a separate task, trained in alphabetical order: bottle, cable, capsule, carpet, grid, hazelnut, leather, metal\_nut, pill, screw, tile, toothbrush, transistor, wood, zipper. Training sets contain only defect-free (nominal) images; test sets contain both nominal and defective images across multiple defect types per category. The discrete category boundaries represent the traditional evaluation paradigm for continual anomaly detection.

\subsection{MVTec-LOCO}
\label{sec:supp_loco}

MVTec-LOCO~\cite{bergmann_beyond_2022} extends MVTec-AD with 5 categories specifically designed to evaluate logical and structural anomalies that cannot be detected by purely local feature comparisons. Categories are trained in alphabetical order. This dataset tests whether methods can detect anomalies requiring global reasoning (\eg, missing or misplaced components) in addition to local texture anomalies.

\subsection{Augmented MTD}
\label{sec:supp_mtd}

The Magnetic Tile Defects (MTD) dataset~\cite{huang_surface_2020} provides a single product category with multiple defect types. To simulate realistic continuous drift scenarios, we augment nominal training images with controlled distortions at 10 ascending intensity levels, each level defining a task. Tasks are trained in ascending order of distortion severity, simulating gradual environmental drift in an industrial setting.

\subsubsection{Color Drift.}
Color distortions are applied by randomly adjusting brightness, contrast, and saturation. Each task defines a narrow, non-overlapping intensity band, ensuring that successive tasks represent progressively stronger---but bounded---distortions without overlap. The 10 intensity bands are $[0.0, 0.05]$, $[0.05, 0.1]$, $[0.1, 0.15]$, $[0.15, 0.2]$, $[0.2, 0.25]$, $[0.25, 0.3]$, $[0.3, 0.35]$, $[0.35, 0.4]$, $[0.4, 0.45]$, $[0.45, 0.5]$. For each training image, a perturbation magnitude is sampled uniformly within the task's band, and applied as a multiplicative factor ($1 \pm v$) to brightness, contrast, and saturation simultaneously, where $v$ is the sampled value. This design ensures each task occupies a distinct region of the distortion space, preventing later tasks from producing near-identity images.

\subsubsection{Blur Drift.}
Gaussian blur is applied with 10 kernel-sigma pairs of increasing severity: $(1, 0.5)$, $(3, 1.0)$, $(5, 1.5)$, $(7, 2.0)$, $(9, 2.5)$, $(11, 3.0)$, $(13, 3.5)$, $(15, 4.0)$, $(17, 4.5)$, $(19, 5.0)$, where the first element is the kernel size and the second is the standard deviation.

\subsubsection{Geometric Drift.}
Geometric distortions combine rotation, translation, scale, and shear transformations applied simultaneously with increasing magnitude. The 10 levels use nominal parameter tuples (rotation${}^\circ$, translation px, scale, shear${}^\circ$): $(2, 1, 0.01, 1)$, $(4, 2, 0.02, 2)$, $(6, 3, 0.03, 3)$, $(8, 4, 0.04, 4)$, $(10, 5, 0.05, 5)$, $(12, 6, 0.06, 6)$, $(14, 7, 0.07, 7)$, $(16, 8, 0.08, 8)$, $(18, 9, 0.09, 9)$, $(20, 10, 0.10, 10)$. Rather than applying these nominal values directly, each task samples its actual parameters from a narrow window \textit{below} the nominal: rotation from $[\text{nom}{-}2, \text{nom}]$, translation from $[\text{nom}{-}1, \text{nom}]$ per axis, scale from $[\text{nom}{-}0.01, \text{nom}]$, and shear from $[\text{nom}{-}1, \text{nom}]$. These parameters are sampled once at task initialization and held fixed for all images within that task, mimicking a stable but shifted operating condition (\eg, gradual conveyor belt misalignment). This narrow-window design ensures each task occupies a bounded region of the geometric distortion space without overlapping previous tasks.

\section{Extended Analysis}
\label{sec:supp_analysis}

\subsection{Detailed Forgetting Analysis}
\label{sec:supp_forgetting}

\subsubsection{MVTec-AD.}
DINOSaur and UCAD exhibit zero forgetting on MVTec-AD (FM\,=\,0.000), though for fundamentally different reasons: DINOSaur's frozen backbone prevents catastrophic forgetting by design, while UCAD's task-specific mechanisms isolate learned distributions. IUF shows strong backward transfer ($-$0.066), suggesting that continued training actually \textit{improves} performance on earlier tasks---though this is less meaningful given IUF's near-chance AUROC. PatchCore's modest forgetting (0.023) demonstrates that experience replay effectively mitigates catastrophic forgetting even without dedicated CL components, particularly when the base method is strong.

\subsubsection{MVTec-LOCO.}
Forgetting on MVTec-LOCO is generally low across all methods, with DINOSaur and UCAD again showing zero forgetting, while PatchCore (0.038) and EfficientAD (0.025) exhibit modest degradation. DNE's negative FM ($-$0.022) indicates backward transfer, though its near-chance AUROC limits the practical significance of this finding.

\subsubsection{Augmented MTD.}
The Forgetting Measures on MTD are uniformly lower than on the discrete-task benchmarks for all methods. On color and blur drift, PatchCore actually exhibits negative FM ($-$0.043 and $-$0.022), indicating backward transfer---continued training on shifted data \textit{improves} detection on earlier tasks. DINOSaur shows near-zero forgetting across all drift types (FM\,$\leq$\,0.013), consistent with its frozen-backbone design.

\subsection{Do CL Components Help? Extended Discussion}
\label{sec:supp_cl_help}

Our most striking finding is that dedicated CL mechanisms do not consistently improve upon traditional AD methods augmented with simple experience replay. On MVTec-AD, DNE (0.500) and IUF (0.508) perform near chance level, while UCAD (0.798) shows meaningful learning but still falls well short of PatchCore with replay (0.925) and DINOSaur (0.968). On the continuous drift protocol, a similar pattern holds: DINOSaur and PatchCore dominate on color and blur drift, and on geometric drift---where all methods struggle---IUF marginally leads (0.525 vs.\ 0.519 for DINOSaur) but all methods remain at chance, so no method meaningfully leads there. On the four protocols with meaningful signal, no dedicated CAD method achieves the top AUROC.

This raises a fundamental question: are the performance gains reported in prior CAD papers genuine algorithmic advances, or artifacts of experimental protocol? Several factors may contribute to this discrepancy:

\begin{itemize}
    \item Prior works may have benefited from extensive hyperparameter tuning on their specific evaluation protocol.
    \item The discrete MVTec-AD category protocol may not be the right testbed for evaluating CL capabilities, as the task shifts are so extreme that even modest CL mechanisms appear effective relative to a non-CL baseline.
    \item Most critically, all of our experiments use the \textit{unsupervised} setting---where only normal samples are available during training---which is substantially harder than the supervised setting used in some prior evaluations of DNE and IUF. The near-chance performance of these methods suggests that their architectures may rely on anomaly examples during training to learn meaningful decision boundaries, a luxury that real industrial deployments rarely afford.
    \item When considering computational constraints for edge deployment, the complexity added by dedicated CL mechanisms provides no clear return on investment.
\end{itemize}

\subsection{Robustness to Drift Type: Extended Analysis}
\label{sec:supp_drift_type}

Comparing the two evaluation paradigms reveals that continuous drift is not uniformly easier than discrete task shifts. Strong methods (DINOSaur, PatchCore) achieve comparable or lower AUROC on MTD color and blur drift than on MVTec-AD, and all methods collapse near chance on geometric drift. The weaker CAD methods (DNE, IUF) show similar near-chance performance regardless of protocol, suggesting their failures are not protocol-specific but rather fundamental limitations of their architectures in the unsupervised setting.

Among drift types, geometric distortion is the most challenging, likely because it fundamentally alters the spatial relationships between patch features. Color and blur distortions modify appearance but preserve spatial structure, making them more amenable to distribution-based detection approaches. This finding has practical implications: industrial scenarios involving mechanical deformation or perspective changes (\eg, conveyor belt misalignment, camera drift) may require fundamentally different representations than those sufficient for photometric variations.

\section{DINOSaur Algorithm}
\label{sec:supp_algorithm}

We provide the complete DINOSaur training and inference procedures in Algorithms~\ref{alg:dinosaur_train} and~\ref{alg:dinosaur_infer}. Notation follows the main text: $\Phi$ denotes the frozen DINOv3 ViT-S/16 encoder, $\mathbf{p}_{i,j}$ a patch feature at spatial location $(i,j)$, $\mathbf{z}$ the CLS token, $\mathcal{M}^{(t)}$ the spatially-indexed memory bank for task $t$, $\mathcal{C}_{i,j}$ the coreset at location $(i,j)$, $\boldsymbol{\mu}^{t}$ the CLS prototype for task $t$, $\mathcal{N}_{i,j}^{r}$ the neighborhood memory within radius $r$, and $\rho$ the coreset ratio.

\begin{algorithm}[tb]
\caption{DINOSaur Training (Task $t$)}
\label{alg:dinosaur_train}
\KwInput{Training set $D_t = \{\mathbf{x}_1, \ldots, \mathbf{x}_{|D_t|}\}$, frozen encoder $\Phi$, coreset ratio $\rho$}
\KwOutput{Memory bank $\mathcal{M}^{(t)}$, CLS prototype $\boldsymbol{\mu}^{t}$}
\BlankLine
\tcc{Feature extraction (Eq.~4)}
\For{each image $\mathbf{x}_k \in D_t$}{
    $\{\mathbf{p}_1^{(k)}, \ldots, \mathbf{p}_N^{(k)}\}, \mathbf{z}^{(k)} \leftarrow \Phi(\mathbf{x}_k)$\;
}
\BlankLine
\tcc{CLS prototype computation}
$\boldsymbol{\mu}^{t} \leftarrow \frac{1}{|D_t|} \sum_{k=1}^{|D_t|} \mathbf{z}^{(k)}$\;
\BlankLine
\tcc{Spatially-indexed coreset selection (Eq.~5)}
\For{each spatial location $(i,j)$ in $\sqrt{N} \times \sqrt{N}$ grid}{
    $\mathcal{P}_{i,j} \leftarrow \{ \mathbf{p}_{i,j}^{(k)} \mid k = 1, \ldots, |D_t| \}$\;
    $M \leftarrow \max(20,\; \lfloor |D_t| \cdot \rho \rfloor)$\;
    $\mathcal{C}_{i,j} \leftarrow \GreedyCoreset{$\mathcal{P}_{i,j}, M$}$\;
}
$\mathcal{M}^{(t)} \leftarrow \{\mathcal{C}_{i,j}\}_{(i,j)}$ \Comment*[r]{Store spatially-indexed memory bank}
\Return{$\mathcal{M}^{(t)}, \boldsymbol{\mu}^{t}$}
\end{algorithm}

\begin{algorithm}[tb]
\caption{DINOSaur Inference}
\label{alg:dinosaur_infer}
\KwInput{Test image $\mathbf{x}_{\text{test}}$, frozen encoder $\Phi$, memory banks $\{\mathcal{M}^{(t)}\}_{t=1}^{T}$, prototypes $\{\boldsymbol{\mu}^{t}\}_{t=1}^{T}$, neighborhood radius $r$}
\KwOutput{Image-level anomaly score $S(\mathbf{x}_{\text{test}})$}
\BlankLine
\tcc{Feature extraction}
$\{\mathbf{p}_1^{\text{test}}, \ldots, \mathbf{p}_N^{\text{test}}\}, \mathbf{z}^{\text{test}} \leftarrow \Phi(\mathbf{x}_{\text{test}})$\;
\BlankLine
\tcc{Task routing via CLS prototypes (Eq.~9)}
$t^* \leftarrow \arg\min_{t} \| \mathbf{z}^{\text{test}} - \boldsymbol{\mu}^{t} \|_2$\;
\BlankLine
\tcc{Neighborhood-restricted anomaly scoring (Eqs.~6--8)}
\For{each spatial location $(i,j)$ in $\sqrt{N} \times \sqrt{N}$ grid}{
    $\mathcal{N}_{i,j}^{r} \leftarrow \bigcup_{\substack{i' \in [i-r, i+r] \\ j' \in [j-r, j+r]}} \mathcal{C}_{i',j'}^{(t^*)}$ \Comment*[r]{Neighborhood memory}
    $s_{i,j} \leftarrow \min_{\mathbf{c} \in \mathcal{N}_{i,j}^{r}} \| \mathbf{p}_{i,j}^{\text{test}} - \mathbf{c} \|_2$ \Comment*[r]{Patch anomaly score}
}
$S(\mathbf{x}_{\text{test}}) \leftarrow \max_{(i,j)} \; s_{i,j}$ \Comment*[r]{Image-level score}
\Return{$S(\mathbf{x}_{\text{test}})$}
\end{algorithm}

\end{document}